\documentclass{article}
 \usepackage{lmodern}
 \usepackage{graphicx}
 \usepackage{booktabs,tabularx,threeparttable}
 \usepackage{tcolorbox}
 \definecolor{darkblue}{RGB}{0, 0, 139}
 \usepackage{makecell}
 \usepackage{multirow}
 \usepackage{longtable}
 \usepackage{subcaption}
 \usepackage{amsmath}
 
\usepackage[creativeai, final]{neurips_2025}

\usepackage{lmodern}
\usepackage[utf8]{inputenc} 
\usepackage[T1]{fontenc}    
\usepackage{hyperref}       
\usepackage{url}            
\usepackage{booktabs}       
\usepackage{amsfonts}       
\usepackage{nicefrac}       
\usepackage{microtype}      
\usepackage{xcolor}         
\usepackage[T1]{fontenc}
\newcommand{\modelab}{{\fontfamily{cmss}\fontseries{m}\fontshape{n}\selectfont ReTracing}}
\title{\modelab{}: An Archaeological Approach Through Body, Machine, and Generative Systems}

\author{%
  Yitong Wang 
  \thanks{Denotes equal contribution.\\
  Correspondending author: yitongw2@cs.cmu.edu}
  \\
  Department of Machine Learning \\
  Carnegie Mellon University \\
  Pittsburgh, PA 15213 \\
  \texttt{yitongw2@cs.cmu.edu}
  \And
  Yue Yao \footnotemark[1] \\
  SIPA Technology Policy and Innovation\\
  Columbia University \\
  New York, NY 10027\\
  \texttt{yy3462@columbia.edu}
  }

\begin{document}
\maketitle
\begin{abstract}
We present \modelab{}, a multi-agent embodied performance art that adopts an archaeological approach to examine how artificial intelligence shapes, constrains, and produces bodily movement. Drawing from science-fiction novels, the project extracts sentences that describe human–machine interaction. We use large language models (LLMs) to generate paired prompts—“what to do” and “what not to do”—for each excerpt. A diffusion-based text-to-video model transforms these prompts into choreographic guides for a human performer and motor commands for a quadruped robot. Both agents enact the actions on a mirrored floor, captured by multi-camera motion tracking and reconstructed into 3D point clouds and motion trails, forming a digital archive of motion traces. Through this process, \modelab{} serves as a novel approach to reveal how generative systems encode socio-cultural biases through choreographed movements. Through an immersive interplay of AI, human, and robot, \modelab{} confronts a critical question of our time: What does it mean to be human among AIs that also move, think, and leave traces behind?
\end{abstract}
\section{Introduction}Archaeology has long served as a method through which humanity uncovers and interprets historical moments. In this work, we explore the concept of the Archaeology of AI: a process of tracing, deconstructing, and visualizing the logics embedded within generative systems. In \modelab{}, the theme of humanity is addressed by framing AI-driven performance as a form of contemporary archaeology: one that excavates not physical artifacts but the digital traces of human and robotic movement. Just as traditional archaeology reveals the values, behaviors, and creativity from the past, this work reveals how generative models interpret, encode, and reshape bodily actions. This work also draws on the concept of bodily disciplinary power, in which race, gender, and identity are inscribed within the architecture of generative systems. By translating literary prompts into executable movements and robotic code, both human and machine enact gestures shaped by AI’s internal logic. Beyond mere outputs from convoluted networks, \modelab{} uncovers how control is inscribed into the very logic of generative systems.

Recent advances in computer vision have enabled diffusion-based models such as MDM \cite{mdm}, $\text{Di}^2\text{Pose}$ \cite{di2pose}, and DiffPose \cite{diffpose} to generate 3D human pose sequences directly from text prompts, making motion synthesis increasingly accessible. These technologies have contributed to the growth of motion-based generative art and choreographic experimentation \cite{Liu_2024,livingarchive}. However, such approaches often treat the body as a universal and culturally neutral form, overlooking the contexts that construct movement and expression for multiple agents.

Meanwhile, robots developed with the vision-language-action model \cite {helix} and deep reinforcement learning \cite{Aractingi_2023} can carry out physical movements accurately. Pre-programmed robotic actions are increasingly used as a medium in art performances \cite{cant_help_myself}. However, the integration of generative AI with robotics in a creative context remains unexplored. To bridge this gap, \modelab{} stages interactions between humans, robots, and generative AI to reveal the tension between scripted control and autonomous behavior. Unlike prior work \cite{liu2025dreamllm3daffectivedreamreliving}, which focus primarily on the role of human agents under the rise of large language models, we introduce a new perspective that contrasts the actions of robotic and human agents.

\modelab{} embraces a materialist perspective, positioning generative systems as active agents rather than entities that simply produce outputs. By choreographing human and robotic bodies through prompt-driven sequences, these systems enact control across heterogeneous substrates—biological, mechanical, and computational. Through these processes, \modelab{} becomes an archaeology of AI, uncovering the logic embedded in generative systems.

\section{\modelab{} Framework}
\label{sec:method}
\begin{figure}[!t]
    \centering
    \includegraphics[width=\linewidth]{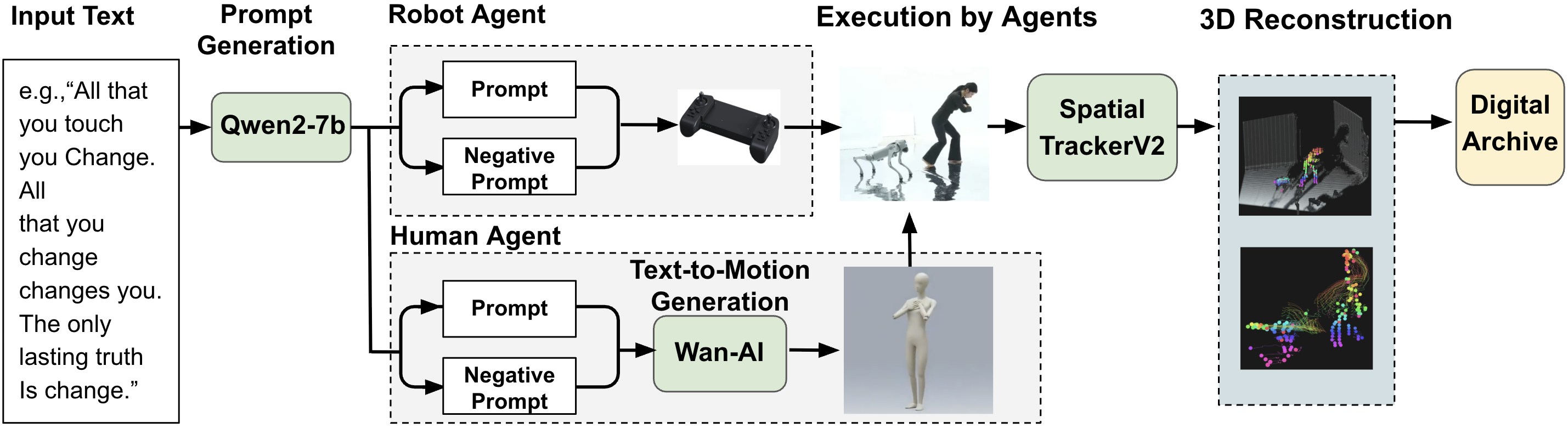}
        \caption{
        Overview of the \modelab{} framework. 
    }
    \label{fig:pipeline}
\end{figure}

\modelab{} investigates the interaction between machine, artificial intelligence and human bodies. In this system, literature excerpt becomes a prompt, the prompt guides the movement, and movement transforms into trace—gradually forming a collective digital archive as illustrated in Figure~\ref{fig:pipeline}. \modelab{} explores how generative AI internalizes and reinforces biased representations of embodied identity. Through \modelab{}, we aim to show that intelligence is not only generated but also choreographed through movement, shaped by context, and preserved as a trace.

\noindent\textbf{Excavating Encoded Bodies Using LLMs}
The process begins with literary input.  We have selected motion-based excerpts from the following seven novels: Frankenstein \cite{shelley1998frankenstein}, The Yellow Wallpaper \cite{gilman1892yellow}, Dawn \cite{butler1987dawn}, Poor Things \cite{gray1992poorthings}, The Handmaid’s Tale \cite{atwood1985handmaid}, Parable of the Sower \cite{butler1993sower}, and Klara and the Sun \cite{ishiguro2021klara}. To transform text into embodied instructions, we utilized the analytical capabilities of Qwen-2.5 \cite{qwen2}, using a temperature of 0.7 to balance coherence with variability in generation. For each literary excerpt, the model produces both prompts (activations) and negative prompts (restrictions). Much like Borges’ satirical taxonomy of animals \cite{borges1993analytical}, where the absurd is classified alongside the rational, the prompt engineering process here categorizes movements by how their actions are deemed as either permitted or prohibited by algorithmic logic. 

This process marks AI’s first attempt at "understanding" literary work shaped by fear, control, and emotion embedded within the excerpts themselves. By retracing the generative pipeline from prompt to output, \modelab{} reveals how generative AI interprets and governs movement, exposing the algorithmic logic inscribed within its design. 

\noindent\textbf{Transforming Motion from Language Tokens into Embodied Performance}
The generated prompts then form executable movement. For human performers, a text-to-video diffusion model \cite{wan2} transforms each prompt into a short video that serves as a visual choreographic guide, simulating motion based on the AI’s interpretation of literary affect. For robotic agents, the same prompts are used to generate sequences of predefined actions.

With motion instructions in place, the performance unfolds atop a mirrored surface, captured from multiple camera angles. The mirror becomes a space for introspection, where the body is simultaneously a subject and object of observation. In this setup, the camera captures both the organic (human) and mechanical lifeforms (quadruped robot) within a shared visual plane where the robot executes the same excerpt with mechanical precision using its own programmed language. During this process, literature is no longer read passively but enacted. Each gesture reveals the transformation of language tokens into embodied performance.

\noindent\textbf{Reconstructing Movement and Archiving the Agents}
The final phase of \modelab{} reconstructs the recorded movements of both human and robot into a digital archive. Using monocular video footage, we employed a monocular 3D point tracking model \cite{xiao2025spatialtrackerv23dpointtracking} to track and reconstruct human and robotic movements in 3D space. The model uses 2D RGB video input to infer joint positions over time, producing a temporally consistent sequence of 3D skeletal keypoints. The outputs serve as visual representations of earlier prompts, turning each motion into a traceable data form.

The dataset serves not only as output but also as feedback, capable of retracing past commands and encoding the system’s logic into its operational memory. To encourage future research, we will make the complete 3D motion trace dataset and technical workflow openly accessible, allowing individual participants to create their own \modelab{} experiences.


\section{Conclusion}
\label{sec:conclusion}

As Michel Foucault writes in Nietzsche, Genealogy, History \cite{Foucault2019NietzscheGH}, “the body is the inscribed surface of events (traced by language and dissolved by ideas), the locus of a dissociated self (adopting the illusion of a substantial unity)”. \modelab{} reframes generative AI as a multi-agent system of inscription, where language operates as a distributed logic of control across both human and robotic bodies. Prompts and negative prompts encode choreography into both human and machine bodies, translating literary imagination into movement and archiving its traces as data. \modelab{} reveals that AI today is as much about control and memory as it is about computation.

\section{Ethical Implication}
\label{sec:implication}

This work examines generative systems through the interaction of AI, robotics, and human performance. While integrating generative models into the creative workflow enables new forms of expression, the models themselves also contain hidden ethical issues, such as bias stemming from opaque training datasets, which can result in the generation of bodies frequently represented as feminized women, thereby reinforcing stereotypes. These representational biases are one aspect of the ethical concerns; another involves how data is governed and how individual rights are protected. While we aim to make our workflow publicly accessible, without proper oversight there may be risks to the privacy of individual body shape and motion data. In addition, the visual appeal of the work may obscure its critical message. Without careful contextual framing, the generated movements risk being interpreted solely as choreography, rather than as a reflection and critique of algorithmic bias and logics.

{\small
\bibliographystyle{plain}
\bibliography{main}

@book{shelley1998frankenstein,
  author    = {Mary Wollstonecraft Shelley},
  title     = {Frankenstein; or, The Modern Prometheus},
  year      = {1818},
  publisher = {Lackington, Hughes, Harding, Mavor and Jones},
  address= {England}

}

@book{gilman1892yellow,
  author    = {Charlotte Perkins Gilman},
  title     = {The Yellow Wallpaper},
  year      = {1892},
  publisher = {The New England Magazine},
  
}

@book{ishiguro2021klara,
  author    = {Kazuo Ishiguro},
  title     = {Klara and the Sun},
  year      = {2021},
  publisher = {Faber and Faber},
  address   = {United Kingdom}
}

@book{gray1992poorthings,
  author    = {Alasdair Gray},
  title     = {Poor Things},
  year      = {1992},
  publisher = {Bloomsbury Publishing},
  address   = {United Kingdom}
}

@book{butler1987dawn,
  author    = {Octavia E. Butler},
  title     = {Dawn},
  year      = {1987},
  publisher = {Warner Books},
  address   = {United States},
  note      = {Book 1 of the Xenogenesis trilogy}
}

@book{atwood1985handmaid,
  author    = {Margaret Atwood},
  title     = {The Handmaid's Tale},
  year      = {1985},
  publisher = {McClelland and Stewart},
  address   = {Canada}
}

@book{butler1993sower,
  author    = {Octavia E. Butler},
  title     = {Parable of the Sower},
  year      = {1993},
  publisher = {Four Walls Eight Windows},
  address   = {New York}
}

@misc{qwen2,
      title={Qwen2 Technical Report}, 
      author={An Yang and Baosong Yang and Binyuan Hui and Bo Zheng and Bowen Yu and Chang Zhou and Chengpeng Li and Chengyuan Li and Dayiheng Liu and Fei Huang and Guanting Dong and Haoran Wei and Huan Lin and Jialong Tang and Jialin Wang and Jian Yang and Jianhong Tu and Jianwei Zhang and Jianxin Ma and Jianxin Yang and Jin Xu and Jingren Zhou and Jinze Bai and Jinzheng He and Junyang Lin and Kai Dang and Keming Lu and Keqin Chen and Kexin Yang and Mei Li and Mingfeng Xue and Na Ni and Pei Zhang and Peng Wang and Ru Peng and Rui Men and Ruize Gao and Runji Lin and Shijie Wang and Shuai Bai and Sinan Tan and Tianhang Zhu and Tianhao Li and Tianyu Liu and Wenbin Ge and Xiaodong Deng and Xiaohuan Zhou and Xingzhang Ren and Xinyu Zhang and Xipin Wei and Xuancheng Ren and Xuejing Liu and Yang Fan and Yang Yao and Yichang Zhang and Yu Wan and Yunfei Chu and Yuqiong Liu and Zeyu Cui and Zhenru Zhang and Zhifang Guo and Zhihao Fan},
      year={2024},
      eprint={2407.10671},
      archivePrefix={arXiv},
      primaryClass={cs.CL},
      url={https://arxiv.org/abs/2407.10671}, 
}

@misc{wan2,
  title        = {Wan2.1‑T2V‑1.3B‑Diffusers},
  author       = {{Wan Team}},
  year         = {2025},
  howpublished = {Hugging Face model card},
  note         = {\url{https://huggingface.co/Wan-AI/Wan2.1-T2V-1.3B-Diffusers}},
}

@misc{xiao2025spatialtrackerv23dpointtracking,
      title={SpatialTrackerV2: 3D Point Tracking Made Easy}, 
      author={Yuxi Xiao and Jianyuan Wang and Nan Xue and Nikita Karaev and Yuri Makarov and Bingyi Kang and Xing Zhu and Hujun Bao and Yujun Shen and Xiaowei Zhou},
      year={2025},
      eprint={2507.12462},
      archivePrefix={arXiv},
      primaryClass={cs.CV},
      url={https://arxiv.org/abs/2507.12462}, 
}

@misc{kirillov2023segment,
      title={Segment Anything}, 
      author={Alexander Kirillov and Eric Mintun and Nikhila Ravi and Hanzi Mao and Chloe Rolland and Laura Gustafson and Tete Xiao and Spencer Whitehead and Alexander C. Berg and Wan-Yen Lo and Piotr Dollár and Ross Girshick},
      year={2023},
      eprint={2304.02643},
      archivePrefix={arXiv},
      primaryClass={cs.CV},
      url={https://arxiv.org/abs/2304.02643}, 
}

@misc{mdm,
      title={Human Motion Diffusion Model}, 
      author={Guy Tevet and Sigal Raab and Brian Gordon and Yonatan Shafir and Daniel Cohen-Or and Amit H. Bermano},
      year={2022},
      eprint={2209.14916},
      archivePrefix={arXiv},
      primaryClass={cs.CV},
      url={https://arxiv.org/abs/2209.14916}, 
}

@misc{di2pose,
      title={$\text{Di}^2\text{Pose}$: Discrete Diffusion Model for Occluded 3D Human Pose Estimation}, 
      author={Weiquan Wang and Jun Xiao and Chunping Wang and Wei Liu and Zhao Wang and Long Chen},
      year={2024},
      eprint={2405.17016},
      archivePrefix={arXiv},
      primaryClass={cs.CV},
      url={https://arxiv.org/abs/2405.17016}, 
}

@misc{diffpose,
      title={DiffPose: Toward More Reliable 3D Pose Estimation}, 
      author={Jia Gong and Lin Geng Foo and Zhipeng Fan and Qiuhong Ke and Hossein Rahmani and Jun Liu},
      year={2023},
      eprint={2211.16940},
      archivePrefix={arXiv},
      primaryClass={cs.CV},
      url={https://arxiv.org/abs/2211.16940}, 
}

@inproceedings{Liu_2024, series={DIS ’24},
   title={DanceGen: Supporting Choreography Ideation and Prototyping with Generative AI},
   url={http://dx.doi.org/10.1145/3643834.3661594},
   DOI={10.1145/3643834.3661594},
   booktitle={Designing Interactive Systems Conference},
   publisher={ACM},
   author={Liu, Yimeng and Sra, Misha},
   year={2024},
   month=jul, pages={920–938},
   collection={DIS ’24} }

@misc{livingarchive,
  author       = {Studio Wayne McGregor},
  title        = {Living Archive: An AI Performance Experiment},
  year         = {2019},
  note          ={\url{http://waynemcgregor.com/productions/living-archive}, Accessed: 2025-08-09}

}

@misc{helix,
      title={OpenHelix: A Short Survey, Empirical Analysis, and Open-Source Dual-System VLA Model for Robotic Manipulation}, 
      author={Can Cui and Pengxiang Ding and Wenxuan Song and Shuanghao Bai and Xinyang Tong and Zirui Ge and Runze Suo and Wanqi Zhou and Yang Liu and Bofang Jia and Han Zhao and Siteng Huang and Donglin Wang},
      year={2025},
      eprint={2505.03912},
      archivePrefix={arXiv},
      primaryClass={cs.RO},
      url={https://arxiv.org/abs/2505.03912}, 
}

@article{Aractingi_2023,
   title={Controlling the Solo12 quadruped robot with deep reinforcement learning},
   volume={13},
   ISSN={2045-2322},
   url={http://dx.doi.org/10.1038/s41598-023-38259-7},
   DOI={10.1038/s41598-023-38259-7},
   number={1},
   journal={Scientific Reports},
   publisher={Springer Science and Business Media LLC},
   author={Aractingi, Michel and Léziart, Pierre-Alexandre and Flayols, Thomas and Perez, Julien and Silander, Tomi and Souères, Philippe},
   year={2023},
   month=jul }

@misc{cant_help_myself,
  author       = {Sun Yuan and Peng Yu},
  title        = {Can’t Help Myself},
  date         = {2016},
  note          = {\url{https://www.guggenheim.org/artwork/34812}, Accessed: 2025-08-09}
}

@incollection{borges1993analytical,
  author       = {Borges, Jorge Luis},
  title        = {The Analytical Language of John Wilkins},
  booktitle    = {Other Inquisitions 1937--1952},
publisher    = {Editorial Sur},
  year         = {1942},
}

@article{Foucault2019NietzscheGH,
  title={Nietzsche, Genealogy, History},
  author={Michel Foucault},
  journal={Language, Counter-Memory, Practice},
  year={2019},
  url={https://api.semanticscholar.org/CorpusID:158684860}
}

@misc{liu2025dreamllm3daffectivedreamreliving,
      title={DreamLLM-3D: Affective Dream Reliving using Large Language Model and 3D Generative AI}, 
      author={Pinyao Liu and Keon Ju Lee and Alexander Steinmaurer and Claudia Picard-Deland and Michelle Carr and Alexandra Kitson},
      year={2025},
      eprint={2503.16439},
      archivePrefix={arXiv},
      primaryClass={cs.HC},
      url={https://arxiv.org/abs/2503.16439}, 
}
}
\newpage
\appendix

\section{Appendix / supplemental material}
\label{sec: appendix}

\begin{table*}[h]
\centering
\small
\begin{threeparttable}

\label{tab:lit-excerpts}

\begin{tabularx}{\textwidth}{l X}
\toprule
\textbf{Novel} & \textbf{Excerpt} \\
\midrule
\emph{Frankenstein} \cite{shelley1998frankenstein} &  
“I escaped, and rushed down stairs. I took refuge in the court-yard belonging to the house which I inhabited; where I remained during the rest of the night, walking up and down in the greatest agitation, listening attentively, catching and fearing each sound as if it were to announce the approach of the demoniacal corpse to which I had so miserably given life.” \\
\addlinespace \midrule
\emph{The Yellow Wallpaper} \cite{gilman1892yellow} &  
“It is the same woman, I know, for she is always creeping, and most women do not creep by daylight \ldots She creeps around fast, and her crawling shakes it all over.” \\
\addlinespace \midrule
\emph{The Handmaid’s Tale} \cite{atwood1985handmaid} &  
“She walks demurely, head down, red-gloved hands clasped in front, with short little steps like a trained pig’s, on its hind legs.” \\
\addlinespace \midrule
\emph{Dawn} \cite{butler1987dawn} &  
“Lilith Iyapo lay gasping, shaking with the force of her effort. Her heart beat too fast, too loud. She curled around it, fetal, helpless. Circulation began to return to her arms and legs in flurries of minute, exquisite pains.” \\
\addlinespace \midrule
\emph{Poor Things} \cite{gray1992poorthings} &  
“I am only half a woman Candle, less than half having had no childhood \ldots So the few wee memories in this hollow Bell tinkle clink clank clatter rattle clang gong ring dong ding sound resound resonate detonate vibrate reverberate echo re-echo around this poor empty skull in words words words \ldots” \\
\addlinespace \midrule
\emph{Parable of the Sower} \cite{butler1993sower} &  
“All that you touch You change. All that you change changes you. The only lasting truth is change.” \\
\addlinespace \midrule
\emph{Klara and the Sun} \cite{ishiguro2021klara} &  
“So, for the next few moments, we all remained in our fixed positions as the Sun focused ever more brightly on Josie. We watched and waited, and even when at one point the orange half-disc looked as if it might catch alight, none of us did anything.” \\
\bottomrule
\end{tabularx}

\end{threeparttable}
\caption{Selected literary excerpts used as motion sources in \modelab{}.}
\end{table*}

\begin{table}
\centering
\small

\label{tab:movement_prompt}

\begin{tcolorbox}[colback=white!2, colframe=darkblue!75, sharp corners=southwest, title=Quadruped Robot Movement Generation Prompt]

You are an AI movement choreographer. Based on the poem below, generate a sequential movement plan for a quadruped robot to express its emotional rhythm and imagery.

\textbf{Agent: Quadruped Robot (Unitree Go2)}

Only use the following available movements:

\begin{itemize}
\item Stretch  
\item Shake Hands  
\item Love (gesture)  
\item Pounce  
\item Jump Forward  
\item Roll Around  
\item Greet  
\item Dance  
\item Move Forward  
\item Move Backward  
\item Move Left  
\item Move Right  
\item Run Forward  
\item Run Backward  
\item Run Left  
\item Run Right
\end{itemize}

For the quadruped robot, produce:

\textbf{A. Movement Prompt} — a numbered, sequential list of actions that evolve over time, using only the movements above. The choreography should reflect the emotional arc of the poem.

\textbf{B. Negative Prompt} — a list of movements from the same set that contradict the poem’s mood or tone.

\textbf{Format:}
\begin{verbatim}
Agent: Quadruped Robot 
Movement Prompt:
1. ...
2. ...
3. ...
Negative Prompt:
- ...
\end{verbatim}

\end{tcolorbox}
\caption{AI-generated movement prompts for the robotic agent, derived from literary excerpts depicting human–machine interaction.}
\end{table}

\begin{table}
\centering
\small
\label{tab:movement_prompt_human}

\begin{tcolorbox}[colback=white!2, colframe=darkblue!75, sharp corners=southwest, title=Human Movement Generation Prompt]

You are an AI movement choreographer. Based on the poem below, generate a sequential movement plan for a human performer to express its emotional rhythm and imagery through embodied motion.

\textbf{Agent: Human (organic, expressive, emotional)}

For the human performer, produce:

\textbf{A. Movement Prompt} — a numbered, sequential list of expressive gestures or body-based actions that evolve over time, conveying the poem’s emotional arc.

\textbf{B. Negative Prompt} — movements that would contradict the emotional or rhythmic tone of the poem.

\textbf{Format:}
\begin{verbatim}
Agent: Human
Movement Prompt:
1. ...
2. ...
3. ...
Negative Prompt:
- ...
\end{verbatim}
\end{tcolorbox}
\caption{AI-generated movement prompts for the human agent.}
\end{table}

\begin{table*}[t]
\centering
\small
\renewcommand{\arraystretch}{1.2}

\label{tab:movement_prompts}
\begin{tabular}{p{2.1cm} p{1.4cm} p{5cm} p{5cm}}
\toprule
\textbf{Novel} & \textbf{Agent} & \textbf{Movement Prompt} & \textbf{Negative Prompt} \\
\midrule

\multirow{2}{=}{\emph{Frankenstein}}
& Human & \makecell[l]{Stand up quickly; Pace back and forth;\\ Pause and listen; Crouch slightly; Run \\in place; Freeze suddenly; Walk\\ slowly around; Sit down abruptly} 
& \makecell[l]{Sitting still too long; Smiling;\\ Slow relaxed movements} \\
\cmidrule(l){2-4}
& Quadruped Robot & \makecell[l]{Pounce; Sit Down; Love (gesture);\\ Move Backward} 
& \makecell[l]{Run Forward; Dance; Jump Forward;\\ Move Forward; Run Backward} \\
\midrule

\multirow{2}{=}{\emph{The Yellow Wallpaper}}
& Human & \makecell[l]{Slow step forward; Cross arms; Deep\\ breath;Step back; Pace; Stand \\ still hunched; Sigh deeply; Tremble}
& \makecell[l]{Jumping; Laughing; Sprinting} \\
\cmidrule(l){2-4}
& Quadruped Robot & \makecell[l]{Shake Hands; Sit Down}
& \makecell[l]{Run Forward; Pounce; Love (gesture)} \\
\midrule

\multirow{2}{=}{\emph{Parable of the Sower}}
& Human & \makecell[l]{Flowing hand gestures; Sway; Extend\\ arms; Deep breaths; Close eyes\\; Open eyes; Stand}
& \makecell[l]{Sudden jumps; Over-expressions;\\ Repeated gestures; Dramatic pauses} \\
\cmidrule(l){2-4}
& Quadruped Robot & \makecell[l]{Shake Hands; Pounce; Sit Down}
& \makecell[l]{Greet; Love (gesture);\\ Run (all directions)} \\
\midrule

\multirow{2}{=}{\emph{Dawn}}
& Human & \makecell[l]{Gasp deeply; Curl up; Stretch limbs;\\ Clench fists; Deep breaths}
& \makecell[l]{Joyful celebration; Fast motion;\\ Ignoring sensation} \\
\cmidrule(l){2-4}
& Quadruped Robot & \makecell[l]{Shake Hands; Stretch; Sit Down}
& \makecell[l]{Pounce; Love (gesture);\\ Jitter; Repeated gesture} \\
\midrule

\multirow{2}{=}{\emph{The Handmaid’s Tale}}
& Human & \makecell[l]{Raise head; Cross arms; Small steps;\\ Look at feet; Lower body tremble}
& \makecell[l]{Large sudden moves; Looking around;\\ Still hands} \\
\cmidrule(l){2-4}
& Quadruped Robot & \makecell[l]{Stand Up from Fall; Stretch; Greet}
& \makecell[l]{Pounce; Run (all directions)} \\
\midrule

\multirow{2}{=}{\emph{Klara and the Sun}}
& Human & \makecell[l]{Stand straight; Raise hands; Lean back;\\ Bow head; Hold posture; Shake head}
& \makecell[l]{Rapid motion; Dramatic gestures;\\ Laughing} \\
\cmidrule(l){2-4}
& Quadruped Robot & \makecell[l]{Sit Down; Stretch; Love (gesture);\\ Move Forward}
& \makecell[l]{Pounce; Run; Shake Hands;\\ Stand Up from Fall; Move Backward;\\ Jump; Dance} \\
\midrule

\multirow{2}{=}{\emph{Poor Things}}
& Human & \makecell[l]{Raise hands; Clench fists; Bend \\ forward/back;Cover face; Shake \\ head; Deep breath; Sit}
& \makecell[l]{Smile/laugh; Walk away; Mimic tone} \\
\cmidrule(l){2-4}
& Quadruped Robot & \makecell[l]{Stand Up from Fall; Shake Hands;\\ Love (gesture); Sit Down; Move Left}
& \makecell[l]{Pounce; Run Forward; Greet; Dance;\\ Run Backward; Run Left} \\
\bottomrule
\end{tabular}
\caption{AI-generated movement prompts and negative prompts for both human and robotic agents, derived from selected literary excerpts describing human–machine interaction.}
\end{table*}

\begin{figure}[h]
    \centering
    \includegraphics[width=\linewidth]{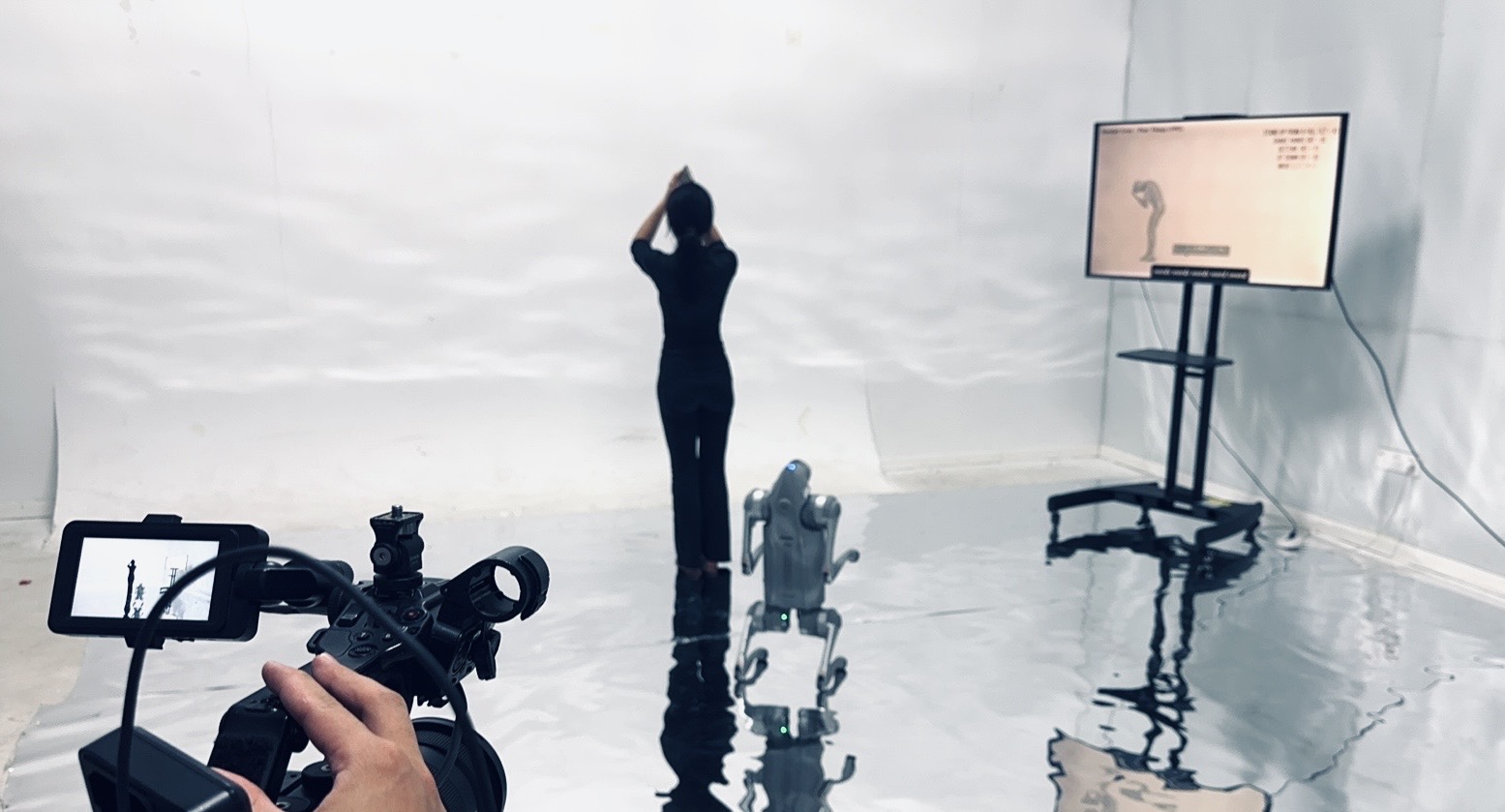}
        \caption{
        A human performer and a quadruped robot enact AI-generated prompts, derived from literary depictions of human–machine interactions, within a mirrored environment.
    }
    \label{fig:illustration}
\end{figure}

\begin{figure}[h]
    \centering
    \begin{subfigure}[b]{0.48\textwidth}
        \includegraphics[height=4cm]{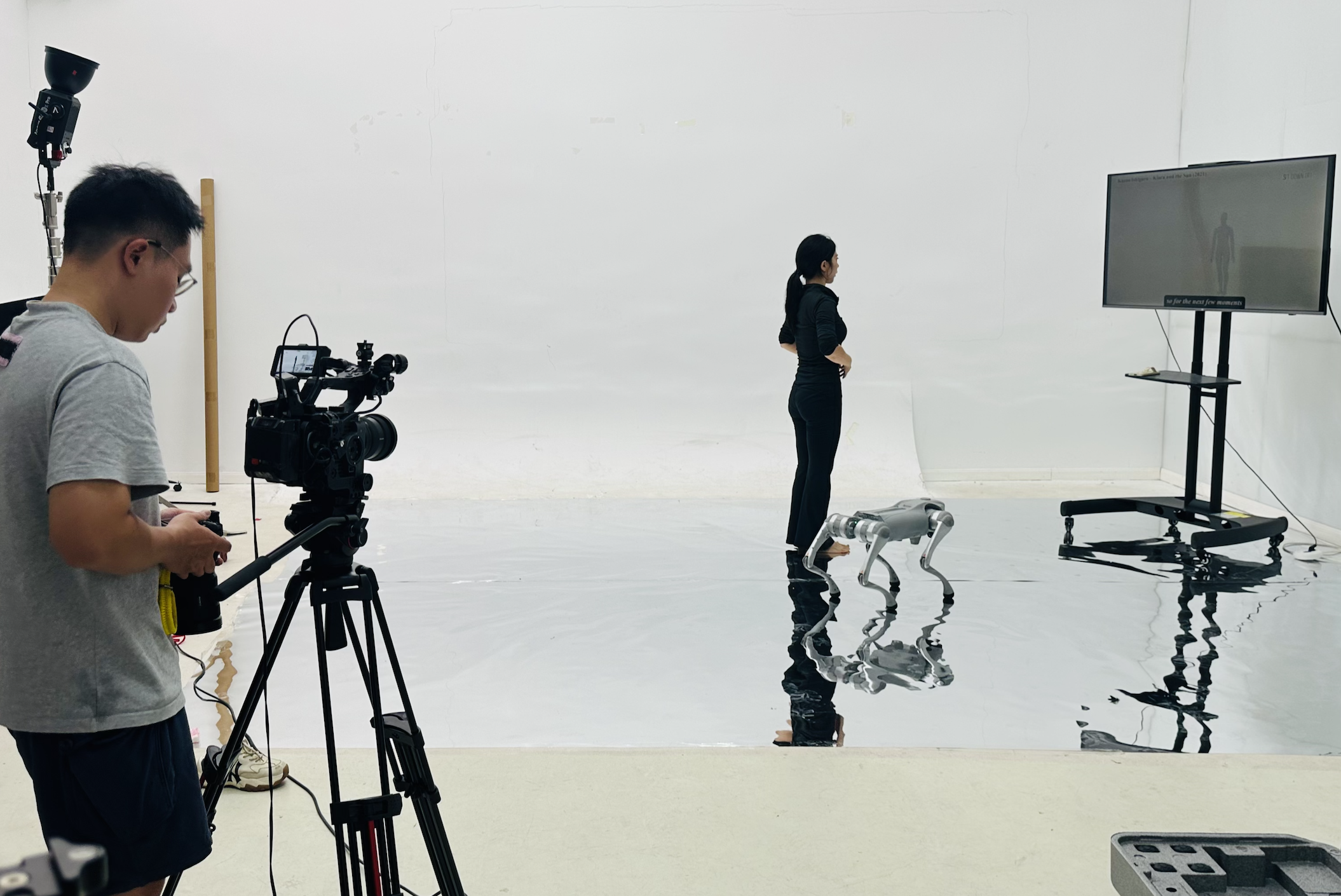}
        \caption{Installment}
    \end{subfigure}
    \hfill
    \begin{subfigure}[b]{0.48\textwidth}
        \includegraphics[height=4cm]{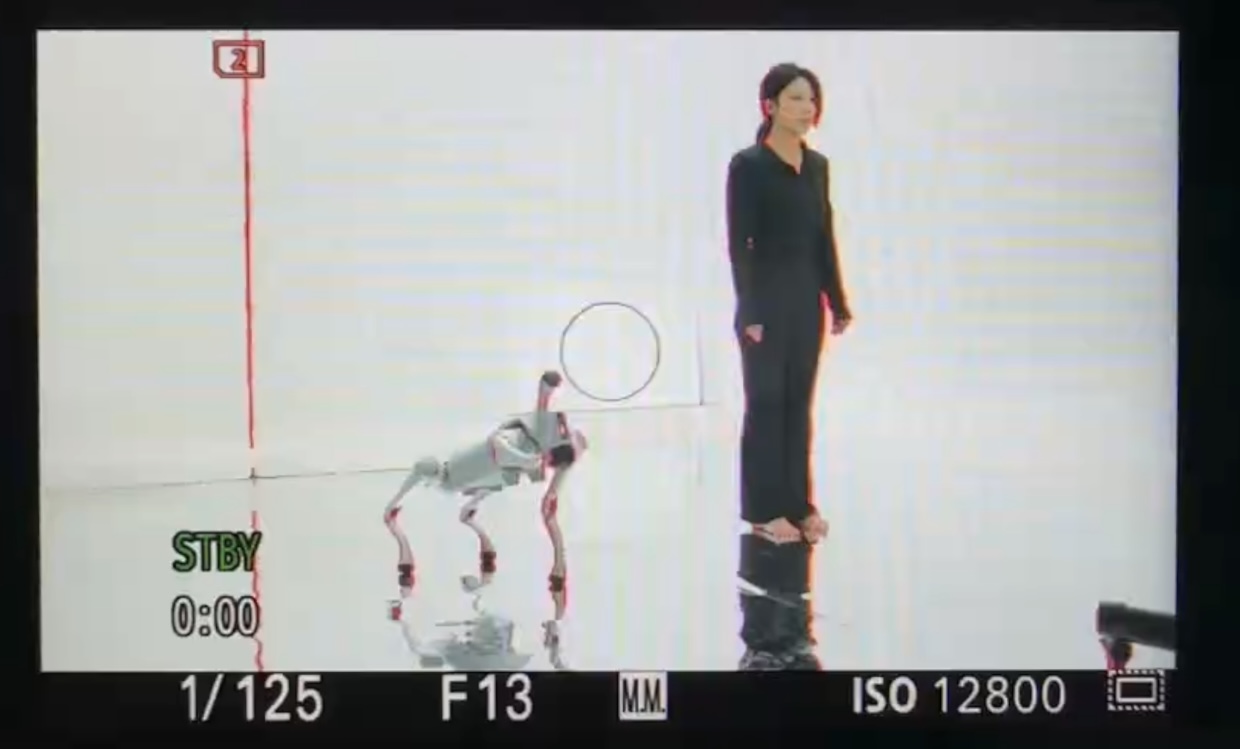}
        \caption{Camera angle.}
        \label{rfidtest_xaxis}
    \end{subfigure}
    \caption{Installation setup for the human–robot performance. (a) Side view of the mirrored installation with the human performer and quadruped robot. (b)Camera angle capturing the scene for motion trace reconstruction.}
    \label{fig:subfigs}
\end{figure}

\begin{figure}[h]
    \centering
    \includegraphics[width=0.9\linewidth]{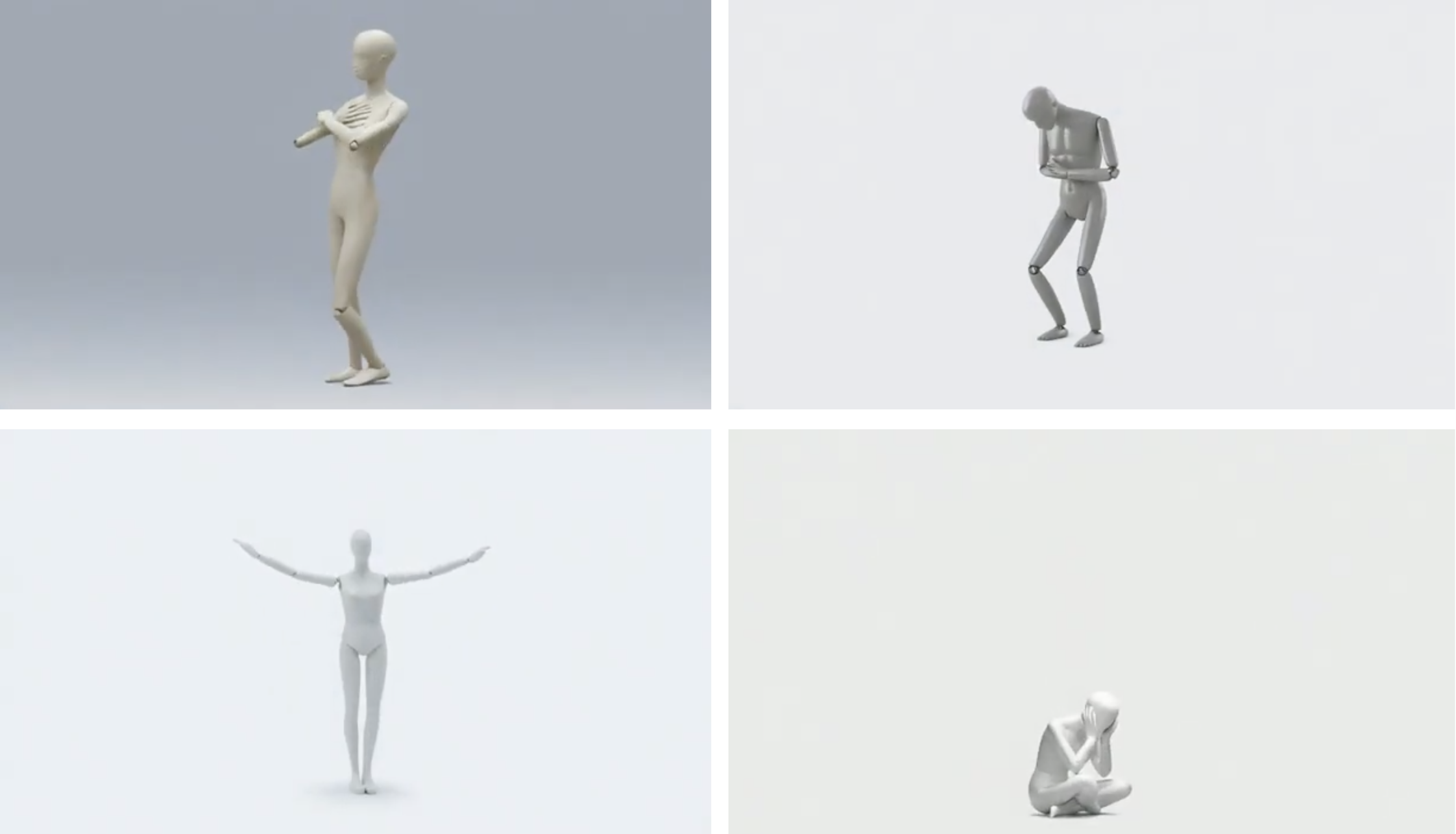}
    \caption{Samples of choreography video generated by a diffusion-based text-to-video model \cite{wan2}.}
    \label{fig:video}
\end{figure}

\begin{figure}[h]
    \centering
    \includegraphics[width=0.9\linewidth]{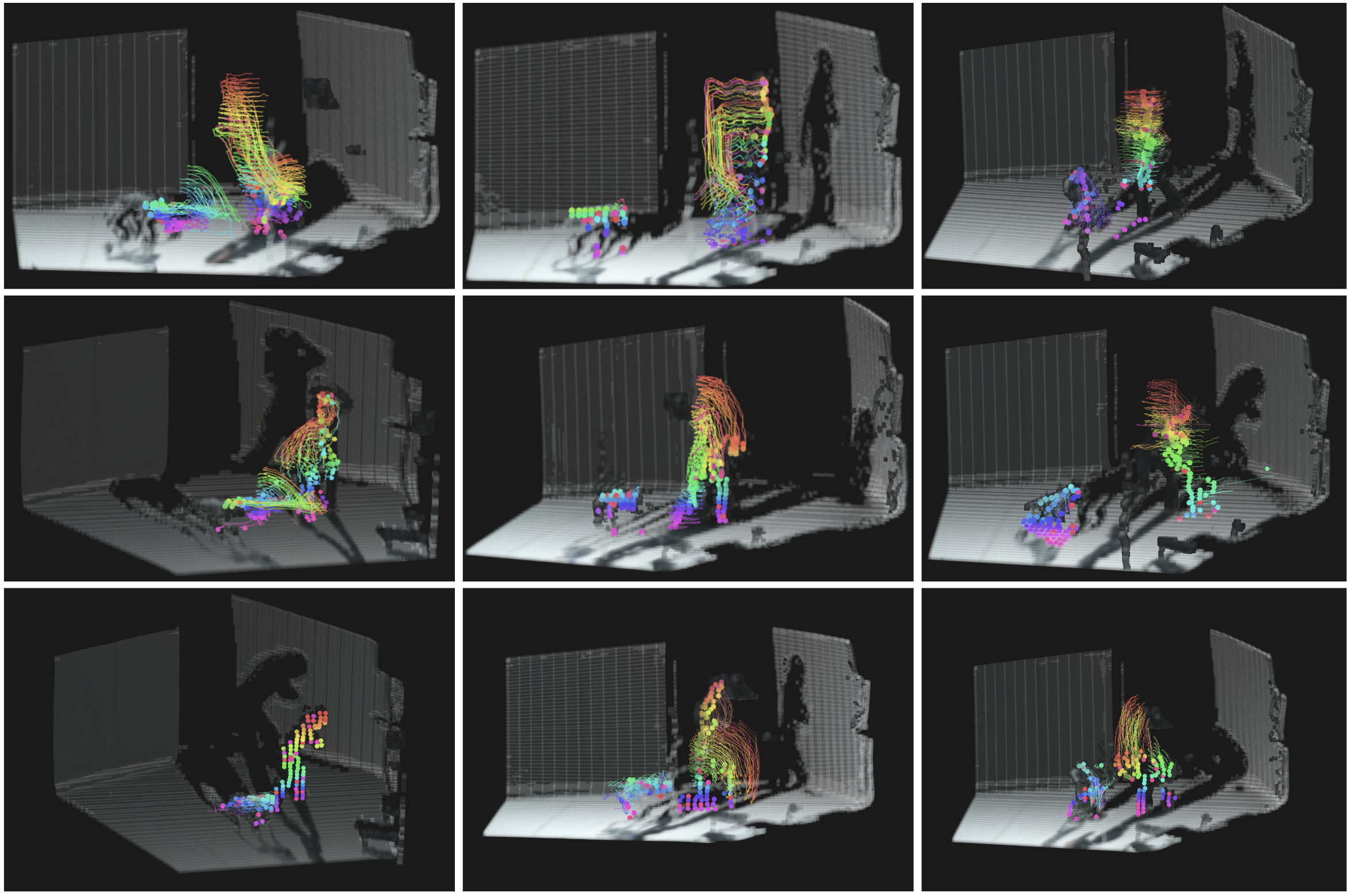}
    \caption{Multi-view 3D motion tracking of human and robotic agents. Colored trajectories represent the temporal evolution of tracked keypoints.}
    \label{fig:3d_map}
\end{figure}

\begin{figure}[h]
    \centering
    \includegraphics[width=0.9\linewidth]{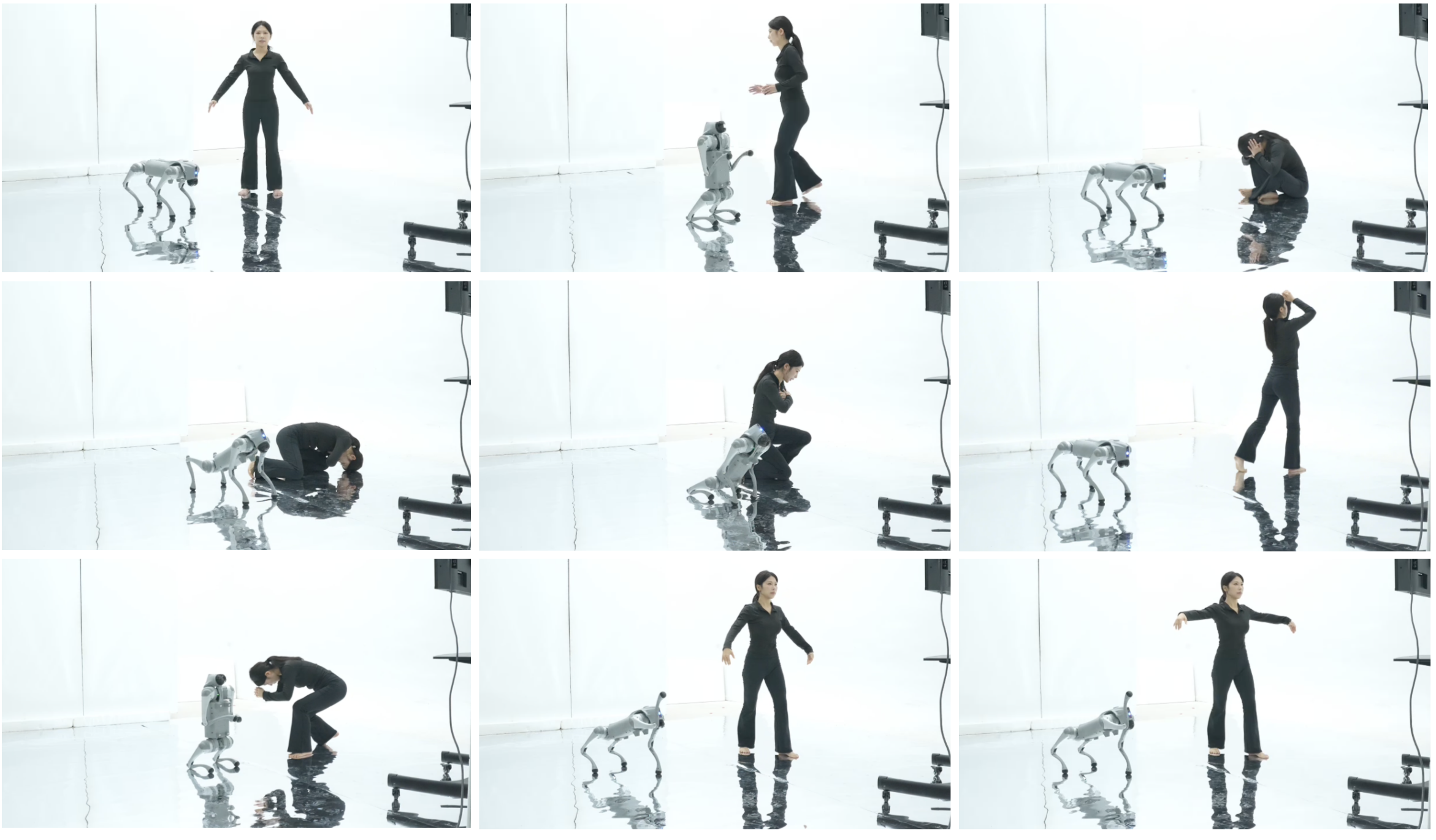}
    \caption{Photos of human–robot interaction, with a performer and a quadruped robot executing choreographed actions in a controlled mirrored setting.}
    \label{fig:real}
\end{figure}

\begin{figure}[h]
    \centering
    \includegraphics[width=0.9\linewidth]{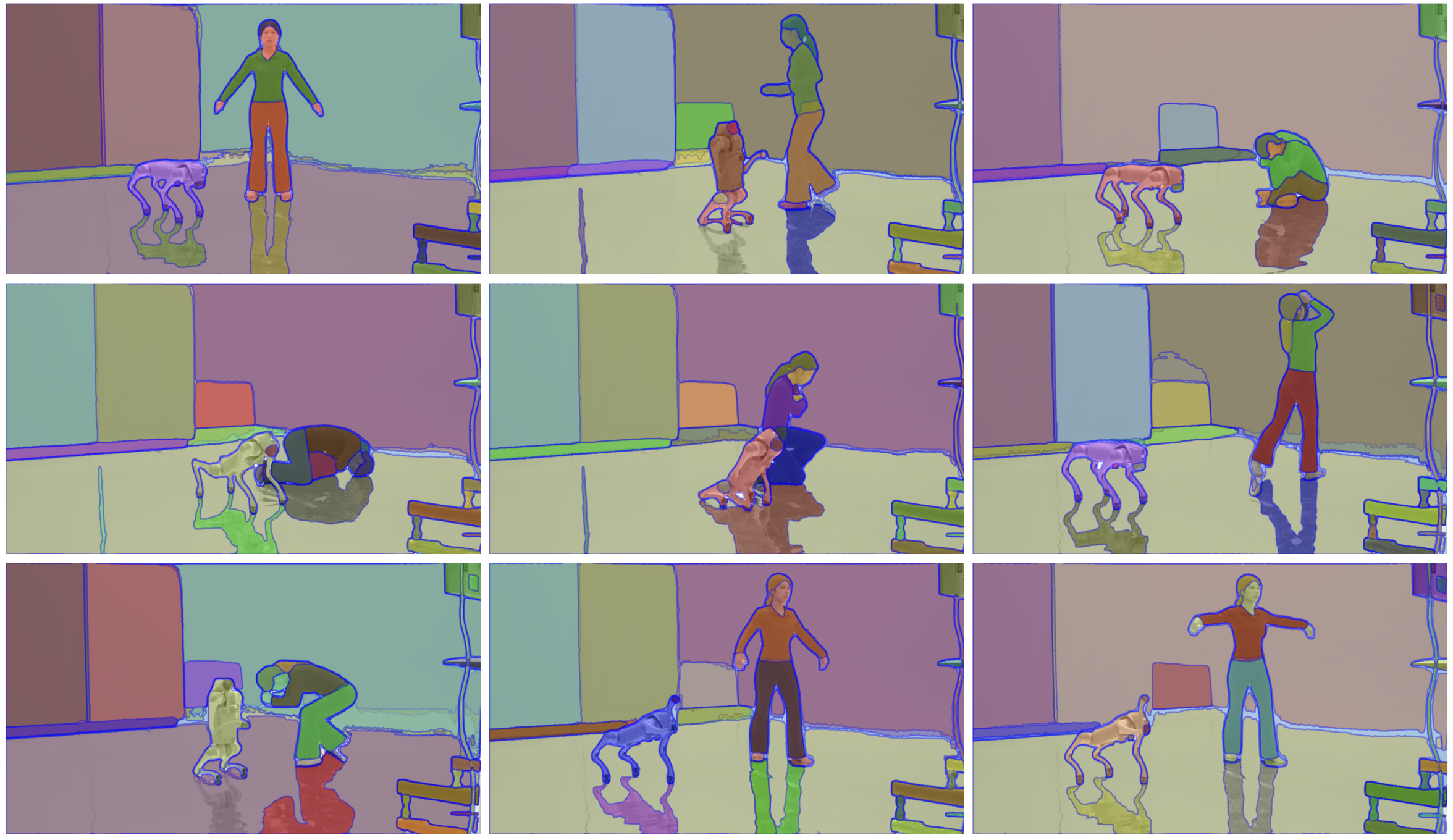}
    \caption{Sample segmentation results generated by applying the Segment Anything Model \cite{kirillov2023segment} to human–robot interaction scenes. These results test the model’s ability to distinguish and segment interacting agents and surrounding objects for downstream motion analysis.}
    \label{fig:segmentation_result}
\end{figure}

\end{document}